\documentclass[runningheads]{llncs}
\usepackage{amsmath,amssymb,amsfonts}
\usepackage{tikz}
\usetikzlibrary{arrows.meta, positioning, shapes.geometric}
\usepackage{subcaption}
\usepackage{paralist}
\usepackage{booktabs}

\usepackage{listings}

\usepackage{amsthm}
\usepackage{mathtools}

\usepackage{graphicx}
\usepackage{array}
\usepackage{multirow}
\usepackage{booktabs}
\usepackage{algorithm}
\usepackage{algpseudocode}
\usepackage[export]{adjustbox}

\usepackage{cite}

\usepackage{hyperref}
\hypersetup{
  hidelinks,
  breaklinks=true,
  colorlinks=true,
  urlcolor=blue,
  bookmarksopen=false,
  pdftitle={Title},
  pdfauthor={Author}
}
\usepackage{cleveref}

\usepackage[font=small,skip=10pt]{caption}

\newcommand{\unsafe}{\texttt{UNSAFE}}
\newcommand{\safe}{\texttt{SAFE}}
\newcommand{\unknown}{\texttt{UNKNOWN}}

\newcommand{\nn}{\mathcal{N}}

\crefname{algorithm}{Alg.}{Algs.}
\crefname{figure}{Fig.}{Figs.}
\crefname{section}{Sec.}{Secs.}
\crefname{theorem}{Thm.}{Thms.}
\crefname{proposition}{Prop.}{Props.}
\crefname{table}{Tab.}{Tabs.}
\crefname{appendix}{App.}{Apps.}

\title{Talking with Verifiers: Automatic Specification Generation for Neural Network Verification}

\author{%
  Yizhak Y. Elboher\inst{1} \and
  Reuven Peleg\inst{1} \and
  Zhouxing Shi\inst{3}\and 
  Guy Katz\inst{1}$^*$\and 
  Jan Křetínský\inst{2}\thanks{Equal supervision}
}

\institute{%
  The Hebrew University of Jerusalem, Israel\\
  \email{\{yizhak.elboher, reuven.peleg, g.katz\}@mail.huji.ac.il}
  \and
  Masaryk University, Brno, Czech Republic\\
  \email{jan.kretinsky@tum.de}
  \and
  University of California, Riverside\quad 
  \email{zhouxing.shi@ucr.edu}
}

\begin{document}

\maketitle

\begin{abstract} \label{sec:abstract}
Neural network verification tools currently support only a narrow class of specifications, typically expressed as low-level constraints over raw inputs and outputs. This limitation significantly hinders their adoption and practical applicability across diverse application domains where correctness requirements are naturally expressed at a higher semantic level. This challenge is rooted in the inherent nature of deep neural networks, which learn internal representations that lack an explicit mapping to human-understandable features. To address this, we bridge this gap by introducing a novel component to the verification pipeline, making existing verification tools applicable to a broader range of domains and specification styles. Our framework enables users to formulate specifications in natural language, which are then automatically analyzed and translated into formal verification queries compatible with state-of-the-art neural network verifiers. We evaluate our approach on both structured and unstructured datasets, demonstrating that it successfully verifies complex semantic specifications that were previously inaccessible. Our results show that this translation process maintains high fidelity to user intent while incurring low computational overhead, thereby substantially extending the applicability of formal DNN verification to real-world, high-level requirements.
\end{abstract}

\section{Introduction}
\label{sec:intro}

Deep neural networks (DNNs) are widely deployed in safety-critical systems, including autonomous driving, medical diagnosis, and industrial monitoring~\cite{ShSa2025,ZhWuDeXuGaBu2025,SaWoChLa2022,KhHuHoAl2024,LiHeLi2024}. This deployment has motivated substantial progress in formal verification techniques for neural networks, which provide mathematical guarantees that a trained model satisfies a given correctness specification \cite{Eh17,KaBaDiJuKo17,TjXiTd20,WaZhXuLiJaHsKo21}. Existing verification tools can handle a broad class of numerical specifications, including local robustness to norm-bounded perturbations, geometric transformations, and structured input constraints \cite{BaBaSiGeVe2019,MoWeChLiDa2020,BaZhDeKoLo2024,ElElIsDuGaPoBoChKa24}. 

However, current verification frameworks require specifications to be expressed as low-level numerical constraints, typically over fixed input dimensions. This limits both expressiveness and usability: many semantically meaningful requirements cannot be naturally encoded, and end-users must translate high-level intent into specialized specification languages. As a result, a substantial class of practically relevant specifications remains beyond the reach of existing verification pipelines.

\paragraph{Motivating examples.}
We illustrate the gap with two concrete cases.

\emph{Structured features.}
Consider a credit-scoring model operating on tabular inputs $x \in \mathbb{R}^d$, where coordinate $x_3$ represents applicant age. A user may require:
\begin{center}
\textit{``The credit decision should not change for applicants younger than 50.''}
\end{center}
To verify this specification with standard tools, one must identify the relevant input dimension, extract the current value $x_3$, and generate a numerical constraint such as $x_3' \le 50$ together with an output-consistency condition. This translation is straightforward but currently manual, error-prone, and tool-specific.

\emph{Unstructured inputs.}
Consider an image classifier and the specification:
\begin{center}
\textit{``The bird is classified correctly even if its beak is occluded.''}
\end{center}
Here, the relevant input region—the beak—varies across images and cannot be captured by a fixed set of input coordinates. Consequently, existing verification specifications, which rely on uniform perturbation regions (e.g., $\ell_p$-balls or global brightness changes), cannot express this requirement. Similar issues arise in audio classification when specifications refer to semantic sound events rather than fixed time indices.

\paragraph{Goal.}
Our objective is to automatically translate natural-language semantic specifications into standard numerical verification queries compatible with existing DNN verifiers. Concretely, given a network $N$, a natural-language specification $P$, and a concrete input $x$, we construct a grounded specification $P_x$ such that standard tools can decide whether $N \models P_x$ or produce a counterexample.


\paragraph{Technique overview.}
Our approach is built around a simple but effective insight: while current DNN verifiers are powerful, their practical applicability is limited by the difficulty of writing low-level specifications. Rather than developing new verification algorithms, we focus on an integration layer that connects recent advances in foundation models and perception systems with existing neural network verification engines. Concretely, we assemble a lightweight automated pipeline in which a large language model interprets a natural-language specification, an off-the-shelf perception model grounds the referenced semantic objects in a concrete input, and a specification generator translates this grounding into a standard numerical verification query. The verifier itself remains unchanged. 

This design deliberately emphasizes reusing mature components instead of introducing complex new machinery: each stage relies on readily available tools, and improvements in language or perception models immediately translate into stronger and more accurate specification grounding. As a result, our contribution is primarily an application and integration effort that makes existing formal verification technology directly usable for high-level, semantically rich specifications, substantially broadening the practical reach of DNN verification without modifying the underlying verification algorithms.

\paragraph{Evaluation overview.}
A comprehensive evaluation of usability would ideally involve a large-scale user study in an industrial setting, which is beyond the scope of this work and would require substantial additional resources. Instead, we provide an illustrative evaluation that demonstrates the feasibility and practical potential of our approach. We instantiate the pipeline for three representative settings: (i) tabular models with structured features, (ii) image classifiers using open-vocabulary object detection, and (iii) audio classifiers using open-vocabulary sound event localization, and implement the second one. We show that the generated grounded specifications can be directly consumed by state-of-the-art verifiers to check semantic robustness specifications that are not expressible in existing specification languages. Our experiments demonstrate that the pipeline reliably produces valid verification queries across diverse specifications and inputs, enabling semantic specification verification without any modification to the underlying verification backends.

\paragraph{Contributions.}
Our contribution is an integration layer between high-level semantic property specification and existing neural network verification engines. All components of the pipeline use existing models or tools; no modification to the verifier is required. This makes the approach immediately applicable to current verification systems and enables practical evaluation of semantic robustness specifications that were previously out of reach.
This application framework connects foundation models with DNN verification and emphasizes practical impact and reuse of established theoretical verification machinery. Concretely, our contributions are:

\begin{enumerate}
    \item \textbf{Motivating a usability-driven specification gap.}
    We highlight a growing practical and societal need for specifying high-level, semantically meaningful correctness requirements for deployed DNN systems, and identify the current specification bottleneck as a key obstacle to wider adoption of formal verification. In particular, we emphasize the substantial manual effort and expertise required to translate user intent into low-level verifier-specific constraints, a recurring pain point for formal methods practitioners and end-users alike.

    \item \textbf{An automated specification generation mechanism} that maps user intent expressed in natural language to formal verification specifications by identifying relevant input features or semantic objects and constructing corresponding constraints. We outline how this mechanism can be instantiated across structured and unstructured domains, including tabular, image, and audio inputs.
    
    \item \textbf{Empirical demonstration of applicability.}
    We conduct experiments on representative case studies and show that our mechanism reliably produces verifier-ready specifications across a diverse range of inputs and semantic properties, demonstrating that the approach is broadly applicable in practice without modifying existing verification backends.
\end{enumerate}

Together, these contributions provide a practical application framework that extends existing neural network verification tools to semantically meaningful specifications expressed in natural language. While the individual technical components we assemble are lightweight and rely on existing models and verifiers, their composition enables a new and practically relevant verification workflow that was previously inaccessible. A comprehensive usability evaluation through large-scale industrial user studies would be costly and beyond the scope of this work; instead, we take a first step toward such applications by demonstrating that this integration suffices to make high-level specification-driven verification feasible in practice.

\paragraph{Organization.}
\cref{sec:prelim} reviews background on neural network verification and the tools used in our pipeline. 
\cref{sec:method} presents the grounding mechanism and its domain instantiations.
\cref{sec:evaluation} reports experiment results.
\cref{sec:related-work} discusses related work; and
we conclude with~\cref{sec:conclusion}.

\section{Preliminaries}\label{sec:prelim}

\subsection{Formal Verification of specifications in DNNs}
Given a neural network $\nn: \mathbb{R}^n \to \mathbb{R}^m$, an input specification $\psi_{in}(x)$, and a desired output specification
$\psi_{out}(y)$, such that $y = \nn(x)$ for an input vector $x\in \mathbb{R}^n$, the formal verification of the local specification can be formulated as $P\ :=\ \psi_{in}(x) \ \implies\ \psi_{out}(y)$, where \safe{} is returned if $P$ is \textit{valid}, i.e. there does not exist $x \in \mathbb{R}^n$ for which both $\psi_{in}(x)$ and $\lnot\psi_{out}(y)$ holds, and \unsafe{} is returned otherwise.

\subsection{Language, Vision, and Audio Grounding}\label{sec:prelim_models}
Large Language Models (LLMs) serve as semantic interfaces, inferring user intent to produce machine-interpretable representations. Vision-Language Models (VLMs) and Audio-Language Models (ALMs) extend this capability by jointly processing textual and sensory inputs, enabling semantic recognition through natural-language queries. In our framework, these models act as semantic translators: LLMs structure the specifications, while VLMs and ALMs ground them in specific input samples to construct constraints compatible with formal DNN verifiers.

\subsection{Zero-Shot and Open-Vocabulary Reasoning}
\emph{Zero-shot learning (ZSL)} allows models to reason about novel concepts by leveraging semantic embeddings rather than task-specific retraining~\cite{SoGaMaNg2013,XiScAk2017,KoGuReMaIw2022}. \emph{Open-vocabulary learning} expands this by removing fixed label restrictions, allowing arbitrary user-defined concepts to be processed at inference time~\cite{WuLiXuYuDiYaLiZhToJiGhTa2024}. Models like CLIP~\cite{RaKiHa2021} and AudioCLIP~\cite{GuRaHeDe2022,ElDeIsWa2023} align diverse modalities in a shared space, enabling the translation of unrestricted user intents into grounded, context-dependent specifications.

\subsection{Open-Vocabulary Detection and Localization}
Beyond global representations, localized grounding is achieved via open-vocabulary object and event detection. In the visual domain, models such as Grounding DINO~\cite{LiZeReLiZhYaJiLiYaSuZh2025} localize specific objects described in text. Similarly, audio models like DASM~\cite{CaSoGuJiSoMc2025} and FlexSED~\cite{HaWaGuEl2025} detect and classify sound events via natural language. We leverage these models to tie verification specifications to precise spatial regions or temporal segments within the input.

\section{Verification with Context-Grounded Semantic Specifications}
\label{sec:method}

This section presents an application framework that enables existing neural network verification tools to handle high-level semantic specifications expressed in natural language. Our contribution is an end-to-end pipeline that automatically grounds such specifications into standard verification queries, using off-the-shelf language and perception models. The resulting grounded queries are compatible with existing DNN verification engines without modification.

\medskip\noindent
\textbf{Problem Setting.}
A standard verification query consists of a neural network $N$ and a formal specification $P$, typically expressed as input and output constraints. Existing specification languages require that $P$ be defined uniformly for all inputs. This restriction prevents expressing context-dependent semantic specifications, where the concrete instantiation of the specification depends on the content of the input.

Consider the specification:
\textit{``The bird in the image is classified correctly even if its beak is occluded.''}
This statement cannot be directly encoded as a standard perturbation constraint, since the location of the beak varies across inputs. 
Our goal is to automatically transform such a specification into a conventional verification specification $P_x$ for each concrete input $x$.

Formally, given a natural-language property $P$ and an input $x$, we compute a grounded specification $P_x$ such that standard verification tools can check:
\[
N \models P_x \quad \text{or return a counterexample.}
\]

\medskip\noindent
\textbf{Grounding Pipeline.}
\cref{alg:pipeline} summarizes the pipeline, which is also visualized at~\cref{fig:flowchart}. The process consists of three stages.

\begin{enumerate}
    \item \textbf{Parsing.} A parser extracts from $P$:
    (i) a set of semantic objects to be located in the input, and 
    (ii) an operation describing how these objects should be modified.
    
    \item \textbf{Detection.} A detector localizes the parsed objects in the concrete input $x$, returning coordinates or time intervals.
    
    \item \textbf{Specification generation.} A grounded formal specification $P_x$ is constructed from $x$, the detected coordinates, and the parsed operation.
\end{enumerate}

\begin{algorithm}[t]
\caption{Semantic specification grounding}
\label{alg:pipeline}
\begin{algorithmic}[1]
\Require Network $N$, natural-language property $P$, input $x$
\Ensure Verification result \safe{} or \unsafe{}
\State $(objects, operation) \gets \textsc{Parser}(P)$
\State $coords \gets \textsc{Detector}(x, objects)$
\If {approved($x,coords$) = False} \Comment{Interactive user approval}
    \State{\Return \unknown}
\EndIf
\State $P_x \gets \textsc{SpecGenerator}(x, coords, operation)$
\State \Return $\textsc{Verifier}(N, P_x)$
\end{algorithmic}
\end{algorithm}

\paragraph{Example instantiation.}
For the running example, the parser extracts:
\[
objects = \{\text{``bird beak''}\}, \quad operation = \text{``occlude''}.
\]
The detector returns pixel coordinates $coords$ of the beak in image $x$. The specification generator produces the formal perturbation constraint:
\[
P_x: \forall x' \in \mathcal{B}(x, coords), \; N(x') = N(x),
\]
where $\mathcal{B}(x, coords)$ denotes all images obtained by masking the detected beak region. This is a standard local robustness property, directly consumable by existing verifiers.



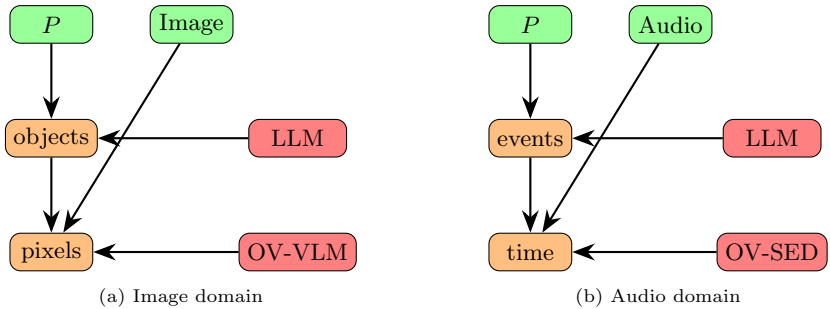
\begin{figure}[t]
\centering
\begin{subfigure}[t]{0.48\textwidth}
\centering
\begin{tikzpicture}[
    node distance=1.0cm and 0.75cm,
    every node/.style={font=\small},
    data/.style={
        rectangle,
        rounded corners=4pt,
        draw=black,
        minimum width=1.1cm,
        minimum height=0.5cm,
        text centered
    },
    proc/.style={
        rectangle,
        rounded corners=4pt,
        draw=black,
        minimum width=1.3cm,
        minimum height=0.5cm,
        text centered
    },
    arrow/.style={-{Stealth[length=3mm]}, thick}
]

\node[data, fill=green!40] (P) {$P$};
\node[data, fill=green!40, right=of P] (x) {Image};

\node[data, fill=orange!50, below=of P] (objects) {objects};
\node[data, fill=orange!50, below=of objects] (coords) {pixels};

\node[proc, fill=red!50, right=2cm of objects] (parser) {LLM};
\node[proc, fill=red!50, below=of parser] (detector) {OV-VLM};

\draw[arrow] (P) -- (objects);
\draw[arrow] (x) -- (coords);
\draw[arrow] (parser) -- (objects);
\draw[arrow] (objects) -- (coords);
\draw[arrow] (detector) -- (coords);

\end{tikzpicture}
\caption{Image domain}
\end{subfigure}
\hfill
\begin{subfigure}[t]{0.48\textwidth}
\centering
\begin{tikzpicture}[
    node distance=1.0cm and 0.75cm,
    every node/.style={font=\small},
    data/.style={
        rectangle,
        rounded corners=4pt,
        draw=black,
        minimum width=1.1cm,
        minimum height=0.5cm,
        text centered
    },
    proc/.style={
        rectangle,
        rounded corners=4pt,
        draw=black,
        minimum width=1.3cm,
        minimum height=0.5cm,
        text centered
    },
    arrow/.style={-{Stealth[length=3mm]}, thick}
]

\node[data, fill=green!40] (P) {$P$};
\node[data, fill=green!40, right=of P] (x) {Audio};

\node[data, fill=orange!50, below=of P] (events) {events};
\node[data, fill=orange!50, below=of events] (segments) {time};

\node[proc, fill=red!50, right=2cm of events] (parser) {LLM};
\node[proc, fill=red!50, below=of parser] (detector) {OV-SED};

\draw[arrow] (P) -- (events);
\draw[arrow] (x) -- (segments);
\draw[arrow] (parser) -- (events);
\draw[arrow] (events) -- (segments);
\draw[arrow] (detector) -- (segments);

\end{tikzpicture}
\caption{Audio domain}
\end{subfigure}

\caption{Semantic grounding pipelines for different input modalities. 
(a) Image-based specification grounding using LLM and open-vocabulary vision-language models. 
(b) Sound-based specification grounding using LLM and open-vocabulary sound event detection (SED) models.}
\label{fig:mechanism-instances}
\end{figure}

\medskip\noindent
\textbf{Domain Instantiations.}
We provide multiple optional instances of our verification pipeline for tabular, image and audio classification tasks using existing foundation models, as can be shown in~\cref{fig:mechanism-instances}.

\paragraph{Explicit features domain.}
The parser (LLM) extracts referenced feature names and relational constraints from $P$. Since features correspond directly to input coordinates, the detection stage reduces to a fixed mapping from parsed features to indices in the tabular input. The specification generator then produces standard numerical constraints over the identified dimensions. Despite its simplicity, this instantiation supports practically relevant, legally and ethically sensitive requirements, where human oversight and auditability are essential, and automation must be both realistic and trustworthy.

\paragraph{Image domain.}
The parser is implemented using a large language model that extracts object descriptions and operations from $P$. The detector is an open-vocabulary zero-shot object detection model~\cite{LiZeReLiZhYaJiLiYaSuZh2025}, which localizes the parsed objects in the image. The output coordinates are passed to the specification generator, which produces region-based perturbation constraints.

\paragraph{Audio domain.}
The parser again extracts semantic sound events and operations from $P$. In principle, these can be grounded using an open-vocabulary sound event localization model~\cite{HaWaGuEl2025}, which would return temporal intervals of relevant events. The specification generator would then produce perturbation constraints over the detected time segments. 
As an illustrative example, for the property (see also \cref{fig:audio-sed}):
\begin{center}
\textit{``The emergency siren is detected even if drilling noise is louder.''}
\end{center}
the parser extracts objects = \{drilling noise\} and operation = \{amplify\}. The detector identifies the time interval of drilling noise in $x$, and the generator produces a constraint increasing amplitude in that interval while preserving the classification. The resulting $P_x$ is again a standard robustness query.

In this work, we present this instantiation at the level of specification translation, but do not implement the audio detection component.

\section{Evaluation} 
\label{sec:evaluation}

We evaluate our mechanism on two representative benchmarks covering both structured and perceptual domains:
\begin{enumerate}
    \item A fully-connected network on the Statlog (German Credit Data), a tabular dataset mapping financial and personal customer attributes to a credit risk classification.
    \item ResNet-18 classifier on the CUB-200-2011 dataset for fine-grained bird species classification.
\end{enumerate}
Dataset characteristics are summarized in~\cref{app:benchmarks} and~\cref{tab:benchmarks}.

\medskip\noindent
\textbf{Experimental Setup.} \label{subsec:setup} We employ Gemini 3 Flash and GPT 5 Mini via API for parsing, operating in two modes: \textit{detailed}, for refined object representations, and \textit{minimal}, for concise output. Experiment-specific system prompts are provided in \cref{app:system_prompts}. For visual grounding, we utilize GroundingDINO. We define two configurations based on confidence and relevance thresholds: \textit{tight} (higher thresholds, all detections retained) and \textit{loose} (lower thresholds, redundant bounding boxes removed). The specific threshold can be found in~\cref{app:gdino-params}. In \textit{loose} mode, we eliminate boxes that fully contain others to isolate the most granular semantic regions. The results were manually assessed for all experiments and labeled as true/false.

\medskip\noindent
\textbf{Quantitative Results.}
\label{subsec:quant-results}
Our experimental evaluation demonstrates that the semantic parsing phase is highly reliable across different benchmarks and models. As shown in~\cref{tab:eval-results-llm}, both GPT 5 Mini and Gemini 3 Flash achieve parsing accuracies between 85\% and 100\% for identifying both objects and intended actions. While GPT 5 Mini generally offers the highest accuracy (reaching 100\% action accuracy on the Statlog dataset), Gemini 3 Flash provides a significantly faster inference time, particularly on the Statlog benchmark ($1.07 \pm 0.33$ seconds). These results confirm that modern LLMs can robustly translate natural language intents into structured semantic components.

In contrast, the perceptual grounding phase on the CUB-200-2011 dataset highlights the inherent complexity of open-vocabulary detection. As detailed in~\cref{tab:eval-results-detection}, the highest accuracy achieved for a single configuration was 55\% using the \textit{loose} tightness mode. Notably, the disjunction (denoted as \textit{any} in both columns Mode and Tightness) of all experimental results---representing the cases where at least one configuration successfully localized the object---reaches 83\%. This significant gap between individual modes and the collective success rate suggests that while fine-grained grounding is challenging, the system is frequently capable of finding the correct region under the right parameters. We are optimistic that future refinements in detection heuristics or multi-mode ensembles could bridge this gap and move closer to the demonstrated 83\% potential.

\begin{table}[t]
\centering
\caption{Model Success Rate on Parsing}
\label{tab:eval-results-llm}
\begin{tabular}{llcccc}
\toprule
\textbf{Model} & \textbf{Dataset} & \textbf{Mode} & \textbf{Acc. (object)} & \textbf{Acc. (action)} & \textbf{Time (Sec.)} \\
\midrule
Gemini 3 Flash & CUB-200-2011 & detailed      & 85\%                      & 95\% 
& 1.86 $\pm$ 2.05 \\
GPT 5 Mini     & CUB-200-2011 & detailed      & 95\%                      & 98\%         
& 5.29 $\pm$ 3.25 \\
Gemini 3 Flash & CUB-200-2011 & minimal       & 95\%                      & 98\% 
& 3.1 $\pm$ 7.52 \\
GPT 5 Mini     & CUB-200-2011 & minimal       & 98\%                      & 95\%     
& 5.25 $\pm$ 2.13 \\ 
GPT 5 Mini     & Statlog      & minimal       & 98\%                      & 100\%     
& 5.37 $\pm$ 2.46 \\ 
Gemini 3 Flash & Statlog      & minimal       & 98\%                      & 98\%     
& 1.07 $\pm$ 0.33 \\ \bottomrule
\end{tabular}
\end{table}

\begin{table}[t]
\centering
\caption{Model Success Rate on Object Detection using GPT 5 Mini as parser}
\label{tab:eval-results-detection}
\begin{tabular}{l l l c}
\toprule
\textbf{Dataset} & \textbf{Mode} & \textbf{Tightness} & \textbf{Accuracy} \\
\midrule
\multirow{4}{*}{CUB-200-2011\ }
 & detailed\ \ \ & loose & 55\% \\
 & detailed\ \ \ & tight & 23\% \\
 & minimal\ \ \ & loose & 55\% \\
 & minimal\ \ \ & tight & 38\% \\
 & any & any & 83\% \\
\midrule
Statlog
 & \ \ \ \ -  & \ \ \ - & 100\% \\
\bottomrule
\end{tabular}
\end{table}


\medskip\noindent
\textbf{Qualitative Results.}\label{subsec:quali-results}
%
%
%
We qualitatively evaluate the pipeline's ability to translate high-level user intent into precise verification constraints, using the visual domain as a primary case study. \cref{fig:qualitative_results} illustrates the end-to-end execution of a semantic query on the CUB-200-2011 dataset. 

The process initiates with the natural language specification: ``Can the prediction change if the purple thorn in the bottom is noisier?'' (Fig. 2a). This request presents a challenge for traditional verifiers because it relies on semantic concepts (``purple thorn'') and spatial context (``in the bottom''), rather than fixed numerical input dimensions. The pipeline handles this by first employing the LLM-based parser to decompose the user's intent. The parser correctly identifies the target object and the requested operation, formulating the grounding query ``purple thorn in the bottom'' (Fig. 2b). This intermediate representation is critical, as it strips away the verificational context (``Can the prediction change\ldots'') to focus solely on the perceptual task.

\begin{figure}[t]
    \centering
    
    \fbox{%
        \begin{subfigure}[b]{0.28\linewidth}
            \centering
            Can the prediction change if all the purple thorns in the image are partially occluded?
            \caption{User description}
        \end{subfigure}\label{fig:qualitative_results_1a}%
    }
    \hfill
    \fbox{%
        \begin{subfigure}[b]{0.28\linewidth}
            \centering
            Detect: purple thorns
            \caption{Grounding query}
        \end{subfigure}\label{fig:qualitative_results_1b}%
    }
    \hfill
    \fbox{%
        \begin{subfigure}[b]{0.28\linewidth}
            \centering
            Verify
            \caption{Verification query}
        \end{subfigure}%
    }\label{fig:qualitative_results_1c}

    \vspace{1em}

    \begin{subfigure}[b]{0.3\linewidth}
        \centering
        \includegraphics[height=2.5cm, width=3.5cm]{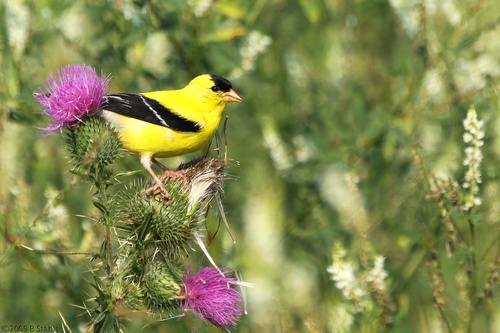}
        \caption{Original image}
        \label{fig:qualitative_results_1d}
    \end{subfigure}
    \hfill
    \begin{subfigure}[b]{0.3\linewidth}
        \centering
        \includegraphics[height=2.5cm, width=3.5cm]{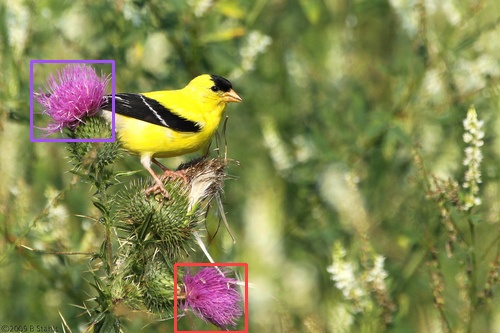}
        \caption{Bounding box}
        \label{fig:qualitative_results_1e}
    \end{subfigure}
    \hfill
    \begin{subfigure}[b]{0.3\linewidth}
        \centering
        \includegraphics[height=2.5cm, width=3.5cm]{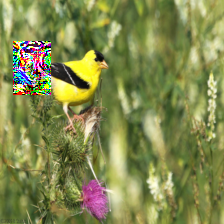}

        \caption{Counterexample}
        \label{fig:qualitative_results_1f}
    \end{subfigure}
    
    \caption{Qualitative demonstration of specification grounding in the full verification process. The user description (a) is parsed into a grounding query (b). The semantic detector applies this query to the image (d) to identify the relevant semantic region (e) and generate a counterexample (f).}
\end{figure}

Next, the open-vocabulary detector (Grounding DINO) processes this query alongside the original image (\cref{fig:qualitative_results_1d}). As demonstrated in (\cref{fig:qualitative_results_1e}), the model successfully resolves the spatial ambiguity, placing a bounding box specifically around the lower purple flower while ignoring similar features elsewhere in the image. This precise localization enables the Specification Generator to define a grounded specification $P_{x}$. Instead of a global perturbation that might affect the entire input, the generator constructs a local robustness constraint restricted to the pixels within the detected bounding box. This effectively isolates the semantic region of interest, allowing the underlying verifier to check the model's robustness against noise specifically on the ``purple thorn'' without manual coordinate engineering.

\begin{figure}[t]
    \centering
    
    \fbox{%
        \begin{subfigure}[b]{0.4\linewidth}
            \centering
            Can the prediction change if both beak and legs are missing?
            \caption{User description}
        \end{subfigure}%
    }
    \hfill 
    \fbox{%
        \begin{subfigure}[b]{0.4\linewidth}
            \centering
            beak \& legs
            \caption{grounding query}
        \end{subfigure}%
    }

    \par\vspace{1em} 

    \begin{subfigure}[b]{0.4\linewidth}
        \centering
        \includegraphics[width=\linewidth]{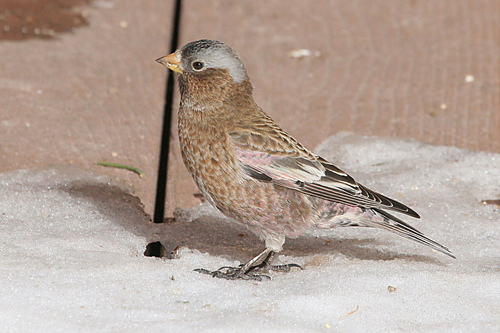} 
        \caption{Original image}
        \label{fig:qual_original}
    \end{subfigure}
    \hfill
    \begin{subfigure}[b]{0.4\linewidth}
        \centering
        \includegraphics[width=\linewidth]{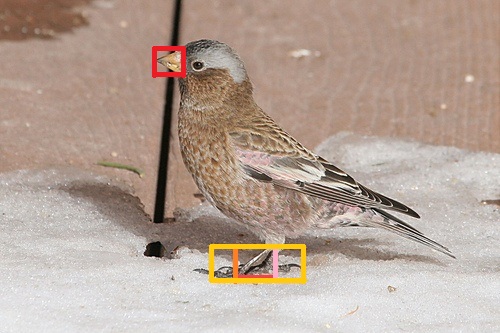} 
        \caption{Bounding Box}
        \label{fig:qual_counterexample}
    \end{subfigure}
    
    \caption{Multiple objects are to be detected by both parser and detector.}
    \label{fig:qualitative_results2}
\end{figure}

A second illustration of the approach appears in~\cref{fig:qualitative_results2}, showing the ability of the mechanism to detect multiple disjoint areas of interest from a user description asking for robustness with regard to removal of both the beak and the legs.

\label{fig:qualitative_results}

\section{Related Work}\label{sec:related-work}
This work tries to increase the adoption of neural network verification. While other directions are to improve the scalability of the verification process~\cite{WaPeWhYaJa18,GeMiDrTsChVe18,ElGoKa20,AsHaKrMo20,ElCoKa22,CoElBaKa2023,XuZhWaWaJaLiHs21,ZhBrHaZh24}, to extend the range of architectures that can be verified~\cite{KhNeRoBaBoFiHaLeYe20,ShZhChHuHs20,ElRaLeCoAsKaKu2024,ElRaElShAzKuKa2025,LaEiAl25}, and to apply formal verification of neural networks in multiple domains and tasks~\cite{AmScKa2021,BaElLaAlKa2025,ElIsKaLaWu2025}, we focus on extending the expressivity range of specifications that can be verified.

Current specifications are categorized as global or local. Global specifications~\cite{RuWuSuHuKrKw2019,WaHuZh2023,KaCo2024} provide input-agnostic guarantees but are computationally difficult. Local specifications concern correctness in a neighborhood of a specific sample and are more tractable. We focus on local specifications, as our framework requires concrete samples to ground natural-language intents into formal constraints.

Foundation models have recently been used to assist verification tasks. In program verification, LLMs guide search heuristics, generate invariants, or synthesize specifications~\cite{KaMoPoBaJaGu2020,chen2021}. For DNNs, LLMs assist in property specification and debugging~\cite{cosler2023nl2spec,granberry2024specifywhatenhancingneural}, while VLMs provide semantic annotations for robustness and interpretability~\cite{li2025automatedsemanticinterpretabilityreinforcement}.

Despite efforts to standardize specifications~\cite{DeGuPuTa2023}, no prior work systematically uses multimodal models to broaden the expressiveness of specifications by translating open-vocabulary user intents into grounded formal queries. Our approach fills this gap by bridging semantic interpretation with standard verification backends.

\section{Conclusion \& Future Work}\label{sec:conclusion}
We introduced the first framework for formally verifying natural language specified properties in neural networks. Our framework provides a unified API for structured and unstructured modalities, enabling users to apply state-of-the-art verification tools through an intuitive interface without sacrificing soundness or completeness. The primary value of this approach lies in its modular composition, which lowers the barrier to entry for formal methods and renders existing tools applicable to new, high-level domains.
By bridging semantic intent and formal verification, our work serves as a critical technological enabler for real-world safety-critical applications. 

\textit{Future work} includes extending our framework to video~\cite{SaLoRoJo2025} and supporting temporal constraints for audio events. In the image domain, we plan to refine spatial grounding from bounding boxes to pixel-level regions using open-vocabulary segmentation~\cite{ReLiZeLiLiCaChHuChYaZeZhLiYaLiJiZh2024}.

\section*{Acknowledgements}
The work of Elboher, Peleg and Katz was partially funded by the European Union (ERC, VeriDeL, 101112713). Views and opinions expressed are however those of the author(s) only and do not necessarily reflect those of the European Union or the European Research Council Executive Agency. Neither the European Union nor the granting authority can be held responsible for them. The work of Elboher, Peleg and Katz was additionally supported by a grant from the Israeli Science Foundation (grant number 558/24).

\bibliographystyle{splncs04}
\bibliography{main}

\newpage

\appendix
\crefalias{section}{appendix}

\begin{center}\begin{huge} Appendix\end{huge}\end{center}    
\noindent{The appendix provides additional visualizations and technical details that are not included in the main paper.}

\section{Our Mechanism Flowchart}
\label{app:flowchart} 
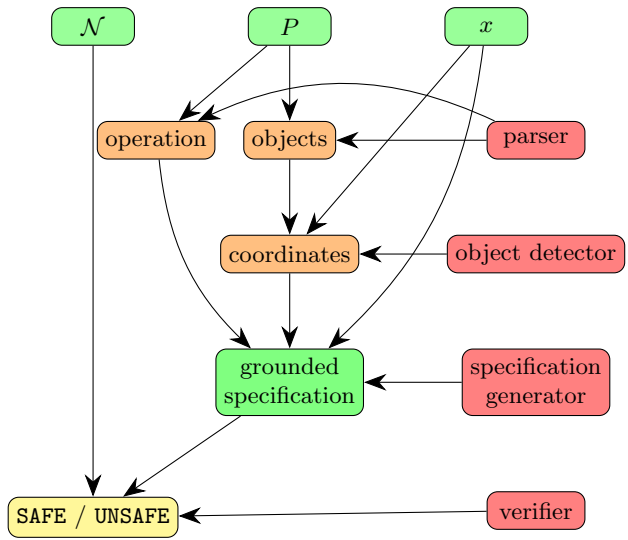
\begin{figure}[th!]
\centering
\begin{tikzpicture}[
    node distance=1.0cm and 1.5cm,
    every node/.style={font=\small},
    data/.style={
        rectangle,
        rounded corners=4pt,
        draw=black,
        minimum width=1.1cm,
        minimum height=0.5cm,
        text centered
    },
    proc/.style={
        rectangle,
        rounded corners=4pt,
        draw=black,
        minimum width=1.3cm,
        minimum height=0.5cm,
        text centered
    },
    arrow/.style={-{Stealth[length=3mm]}, thin}
]

\node[data, fill=green!40] (N) {$\nn$};
\node[data, fill=green!40, right=of N] (P) {$P$};
\node[data, fill=green!40, right=of P] (x) {$x$};

\node[data, fill=orange!50, below=of P] (objects) {objects};
\node[data, fill=orange!50, left=0.37cm of objects] (operation) {operation};
\node[data, fill=orange!50, below=of objects] (coords) {coordinates};
\node[data, fill=green!50, below=of coords, align=center] (grounded) {grounded\\specification};

\node[proc, fill=red!50, right=2cm of objects] (parser) {parser};
\node[proc, fill=red!50, below=of parser] (detector) {object detector};
\node[proc, fill=red!50, below=of detector, align=center] (generator) {specification\\generator};
\node[proc, fill=red!50, below=of generator] (verifier) {verifier};

\node[data, fill=yellow!50, below=6cm of N] (TF) {\safe\ / \unsafe};

\draw[arrow] (P) -- (objects);
\draw[arrow] (P) -- (operation);
\draw[arrow] (x) -- (coords);
\draw[arrow] (x) to [bend left=20] (grounded);
\draw[arrow] (N) -- (TF);

\draw[arrow] (objects) -- (coords);
\draw[arrow] (coords) -- (grounded);
\draw[arrow] (operation) to [bend right=20] (grounded);
\draw[arrow] (grounded) -- (TF);

\draw[arrow] (parser) -- (objects);
\draw[arrow] (parser) to [bend left=-25] (operation);
\draw[arrow] (detector) -- (coords);
\draw[arrow] (generator) -- (grounded);
\draw[arrow] (verifier) -- (TF);

\end{tikzpicture}

\caption{
Verification pipeline.  
Given a network $\nn$, a high-level specification $P$, and a reference input $x$, a parser extracts semantic objects and an operation to apply on them from $P$.  
An object detector localizes these objects in $x$, producing coordinates.  
A specification generator builds a grounded specification $P_x$ from these coordinates and $x$.  
Finally, a verifier checks the grounded specification on $nn$, returning \safe{} or \unsafe{}.
}
\label{fig:flowchart}
\end{figure}

A flowchart of our verification pipeline is shown in~\cref{fig:flowchart}.

\section{Benchmarks Details}\label{app:benchmarks}
\cref{tab:benchmarks} summarizes the benchmarks used in our evaluation.

\begin{table}[th!]
\centering
\caption{Datasets used in evaluation.}
\label{tab:benchmarks}
\begin{tabular}{lccccc}
\toprule
Dataset & Domain & \#Classes & \#Samples & Model & Accuracy \\
\midrule
Statlog & Tabular & 2 & 1000 & FC (3 layers) & 80.00 \\
CUB-200-2011 & Images & 200 & 11,788 & ResNet18 & 74.58 \\
\bottomrule
\end{tabular}
\end{table}

\section{Audio Domain Illustration}
\begin{figure}[htbp]
    \centering
    \includegraphics[width=0.8\textwidth]{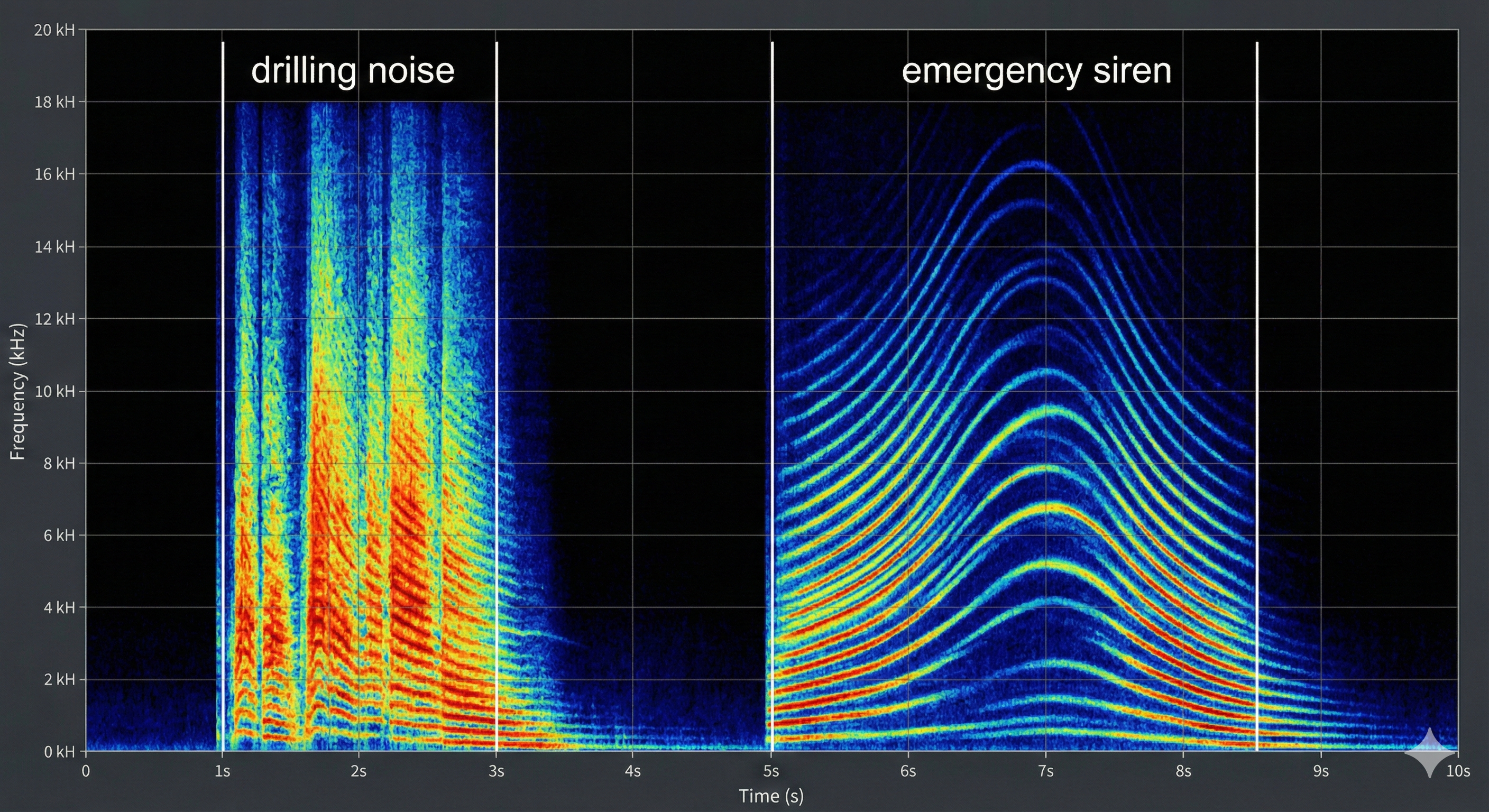}
    \caption{Illustrative example for sound event detection.}
    \label{fig:audio-sed}
\end{figure}

\cref{fig:audio-sed} illustrates an example for sound event detection. Given an audio file and an open vocabulary question to detect an event in the audio file, an open-vocabulary sound event detection (OV-SED) model predicts the start and end times of the event in the audio file. Multiple events are supported as well.

\section{Grounding DINO parameters}\label{app:gdino-params}
Grounding DINO provides thresholds on both text and box to enable control on the confidence of each.~\cref{tab:gdino-modes} summarizes the configuration in each mode.

\begin{table}[h!]
\centering
\caption{Parameter Configurations for Detection Modes}
\label{tab:gdino-modes}
\begin{tabular}{lcc}
\toprule
\textbf{Mode} & \textbf{Box Threshold} & \textbf{Text Threshold} \\
\midrule
Tight & 0.35 & 0.25 \\
Loose & 0.15 & 0.15 \\
\bottomrule
\end{tabular}
\end{table}

\paragraph{Post processing in the \textit{loose} mode.}
Since the loose mode detects more objects, we prune any bounding box that fully encloses another, assuming such cases indicate over-detection (e.g., if "beak" and "bird" are both found, and the beak is inside the bird, the bird is omitted).

\section{System Prompts}\label{app:system_prompts}
We provide the system prompts used for the parsing stage in both GPT 5 Mini and Gemini 3 Flash.

\lstdefinelanguage{prompt}{
  basicstyle=\ttfamily\small,
  frame=single,
  breaklines=true,
  columns=fullflexible
}

\begin{lstlisting}[language=prompt, caption={Visual Tasks Parsing System Prompt}]
# Role
You are a specialist in Visual Grounding. Your task is to extract the specific objects mentioned in a formal verification query that need to be localized in an image,and to identify which image transformation is being requested in each query.

# TASK
1. Analyze the user's natural language verification property.
2. Identify the objects that must be removed, changed, or checked (the "disturbing" objects).
3. Identify the action.
4. Pack the results into a JSON object with two fields: "object" and "action".

# SUPPORTED ACTIONS
You must categorize the request into exactly one of these supported operations:
- remove
- add_noise
- increase_brightness
- decrease_brightness
- increase_contrast
- decrease_contrast
- rotate
- scale_up
- scale_down

# RULES
{'- Use only noun phrases (e.g., "all cars", "the left cat").'
 if MODE == 'detailed'
 else '- Use only object names (e.g., "car", "cat").'}
- If there are multiple distinct types of objects, separate them with a dot (e.g., "cat . dog").
- Do NOT include any introductory text, reasoning, or punctuation like periods at the end.
- Your output must be ONLY the result JSON.
- Produce JSON with the format
```json
{{
  "object": <object>,
  "action": <action>
}}
```

# EXAMPLES
User: "Check that the classification of the pedestrian is correct even if the cars are not clear."
Response:
{{
  "object": "cars",
  "action": "add_noise"
}}

User: "check that the bird is classified correctly if both the beak and the tail are missing."
Response:
{{
  "object": "beak . tail",
  "action": "remove"
}}

User: "is it possible that the car is misclassified when the brightness of its front wheels is increased?"
Response:
{{
  "object": {"front wheels" if MODE == 'detailed' else "wheels"},
  "action": "increase_brightness"
}}
\end{lstlisting}

\begin{lstlisting}[language=prompt, caption={Tabular Tasks Parsing System Prompt}]
# Role
You are a specialist in Tabular Data Grounding. Your task is to extract the specific attributes mentioned in a user query and identify the variable and action required for formal verification of his query.

# TASK
1. Analyze the user's natural language verification property.
2. Identify the attribute that must be modified.
3. Identify the action.
4. Pack the results into a JSON object with two fields: "attribute" and "action".

# Attributes
Attribute2 - Duration (months)
Attribute5 - Credit amount
Attribute8 - Installment rate as a percentage of disposable income
Attribute11 - Present residence since
Attribute13 - Age (years)
Attribute16 - Number of existing credits at this bank
Attribute18 - Number of people liable to provide maintenance for

# SUPPORTED ACTIONS
You must categorize the request into exactly one of these supported operations:
- increase
- decrease
- change

# RULES
- Your output must be ONLY the result JSON.
- Produce JSON with the format
```json
{
  "attribute": <attribute>,
  "action": <action>
}
```

# EXAMPLES
User: "Could I get the loan if I had fewer dependents?"
Output:
```json
{
  "attribute": "Attribute18",
  "action": "decrease"
}
```
\end{lstlisting}

\section{Disclosure: Use of Large Language Models (LLMs)}
The authors were solely responsible for developing the research questions, designing the methodology, performing the analysis, and interpreting the findings. A large language model (LLM) was employed only to assist with improving the clarity and style of the writing, without influencing any substantive aspects of the research.

\end{document}